\documentclass[letterpaper, 10 pt, conference]{ieeeconf}  

\IEEEoverridecommandlockouts                              

\usepackage{graphicx} 
\usepackage[dvipsnames]{xcolor}
\usepackage{algorithm}
\usepackage{algpseudocode}
\usepackage{booktabs} 
\usepackage[font=small,belowskip=-1.5pt]{caption} 
\makeatletter
\let\NAT@parse\undefined
\makeatother
\usepackage{hyperref}
\hypersetup{
    colorlinks=true,
    linkcolor=blue,
    filecolor=magenta,      
    urlcolor=cyan,
    pdftitle={DroNERF},
    pdfpagemode=FullScreen,
    }

\renewcommand{\baselinestretch}{0.9}
\renewcommand{\baselinestretch}{0.92}
\title{\LARGE \bf
\textit{DroNeRF}: Real-time Multi-agent Drone Pose Optimization \\for Computing Neural Radiance Fields}
\author{Dipam Patel, Phu Pham and Aniket Bera \\
{Department of Computer Science, Purdue University, USA}\\
Supplemental Version: {\small \href{https://ideas.cs.purdue.edu/research/dronerf}{ideas.cs.purdue.edu/research/dronerf}}
\vspace{-10pt}
}

\begin{document}

\maketitle
\thispagestyle{empty}
\pagestyle{empty}

\begin{abstract}

We present a novel optimization algorithm called \textit{DroNeRF} for the autonomous positioning of monocular camera drones around an object for real-time 3D reconstruction using only a few images. Neural Radiance Fields, or NeRF, is a novel view synthesis technique used to generate new views of an object or scene from a set of input images. Using drones in conjunction with NeRF provides a unique and dynamic way to generate novel views of a scene, especially with limited scene capabilities of restricted movements. Our approach focuses on calculating optimized pose for individual drones while solely depending on the object geometry without using any external localization system. The unique camera positioning during the data capturing phase significantly impacts the quality of the 3D model. To evaluate the quality of our generated novel views, we compute different perceptual metrics like the Peak Signal-to-Noise Ratio (PSNR) and Structural Similarity Index Measure (SSIM). Our work demonstrates the benefit of using an optimal placement of various drones with limited mobility to generate perceptually better results. 

\end{abstract}

\section{INTRODUCTION}

The camera's location is of utmost importance for capturing any scene since it directly affects the performance of the vision system involved in interpreting visual data. Especially for a task like 3D reconstruction, the camera pose and exposure of the subject under consideration significantly impact the quality of the final reconstructed model. When the system has to incorporate a random camera movement, unusual and misaligned from everyday scenarios, it results in poor reconstruction or failure. At that point, the system faces unpredictable changes in the scene, which makes the reconstruction challenging \cite{Gonzalez2009}. Mainly for applications involving scene understanding in challenging scenes, using drones for novel view synthesis could provide a unique and dynamic way to generate new views of a scene or object, opening up possibilities for new applications in areas such as virtual reality, gaming, and entertainment. The quality of the NeRF model strongly depends on the camera positions of the images (especially if the number of images is sparse, i.e., less than ten images). Our paper focuses on the problem of specific camera placement for a swarm of monocular camera drones. We aim to determine an independent optimal pose for each drone to capture the least number of images based on the object's geometry.

\begin{figure}[h]
  \centering
  \includegraphics[width=1\linewidth]{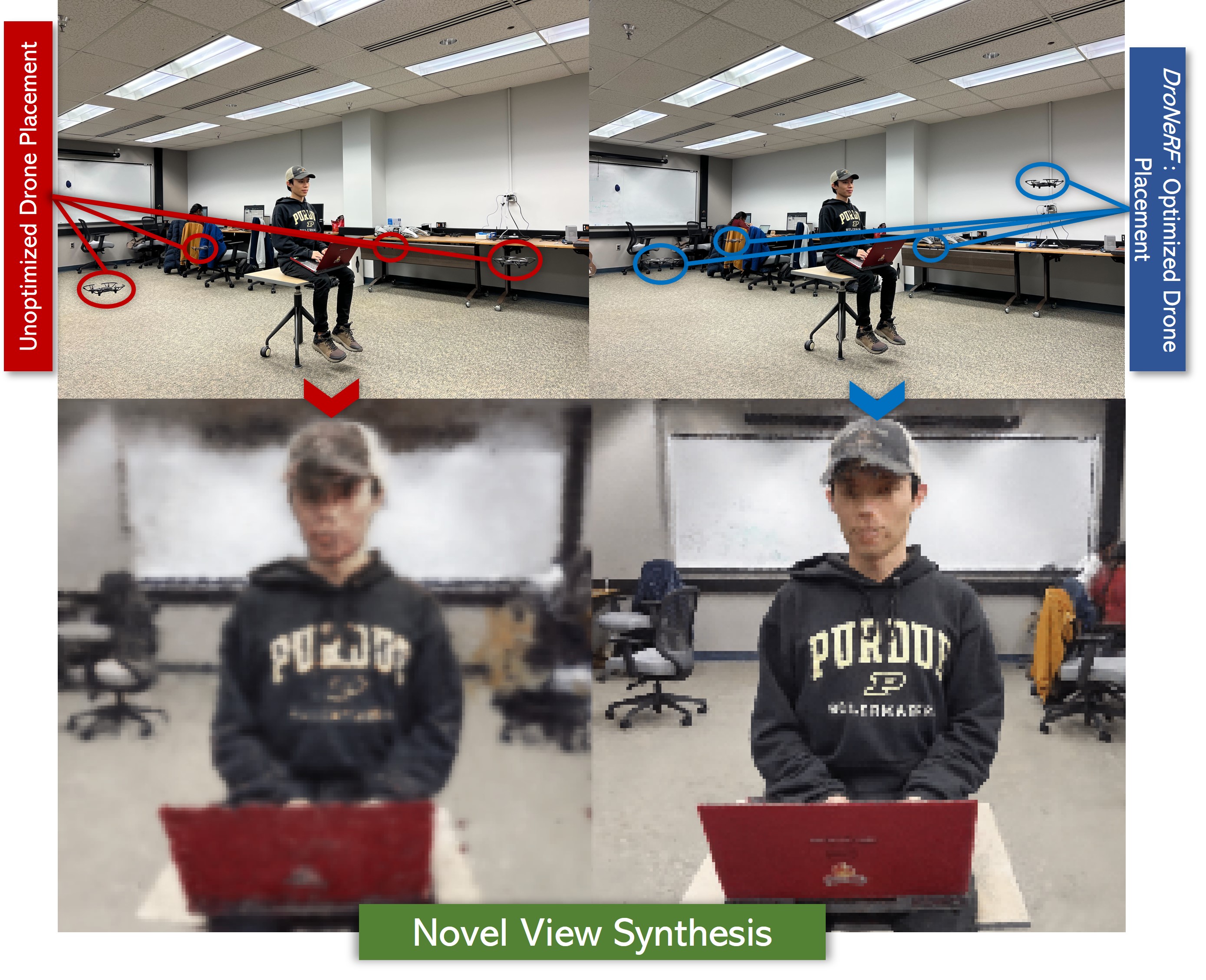}
  \caption{\textit{Comparison of 3D reconstructions with different flight configurations - The left half shows the drones in a conventional circular pattern with no priority to object geometry. The right half shows the optimized position of the drones with a focus on the top half of the person. As a result, the drone facing the person has a higher height compared to the rest of them.}}
  \label{fig:dronemain}
\end{figure}

Our algorithm prioritizes covering the critical aspect of the target since that makes the reconstruction extremely useful. For instance, while capturing images of a person eating food, the face of the person, the location of the person’s hands, and the food on the table are more important than other parts of the table or the human body. Especially for concave objects, more focus must be closer to the concavity to optimally capture more of the geometry. This paper showcases how the target's orientation, shape, and size result in drone pose adjustments around it \cite{Chabra2017}.

On the other hand, Neural Radiance Field (NeRF) \cite{NeRF} has recently enabled novel 3D reconstruction based on 2D images, producing highly realistic and accurate results. It is advantageous for rendering high-resolution photorealistic novel views of real objects. These 3D reconstructions usually use expensive equipment like RGB-D cameras or Lidar sensors. However, the former can fail under direct sunlight, while the latter remains heavier and more expensive. The simple, inexpensive, and lightweight alternative is using a monocular camera, which would be a more convenient solution.

M\"uller et al. \cite{instant-ngp} fit a radiance field in real-time given only posed images. Chng et al. \cite{GARF} and Lin et al. \cite{BARF} unveiled that ground truth poses are not compulsory if sufficiently good initial estimates are available. They build accurate radiance fields from a stream of consecutive images without requiring ground truth poses or depth maps as input. 

When deciding to capture object photos in the real world, in most state-of-the-art applications in the computer vision community \cite{Gonzalez2009}, most camera placements are placed symmetrically across a virtual hemisphere around the object. This camera distribution generally works well but assumes that the objects are relatively convex and evenly spread out. It fails when objects are oddly shaped, non-symmetric, or objects with concavities.

Our approach is an optimization technique to iteratively understand the object geometry and re-align the drone/cameras to a position where more mesh details are captured (i.e., more mesh triangles are visible to the camera).

Since this work does not focus on autonomously finding the subject of interest in the environment, this initial placement helps prioritize the subject from the point-of-view of finding the optimal camera pose compared to conventional positions. Our system is inspired by the work on the region detection strategy by Kong et al. \cite{Kong2013} and the surface orientation estimation by Chabra et al. \cite{Chabra2017}. Figure \ref{fig:dronemain} directly compares different flight configurations and their respective 3D reconstructions.

\begin{figure}[h]
  \centering
  {\includegraphics[width=1\linewidth]{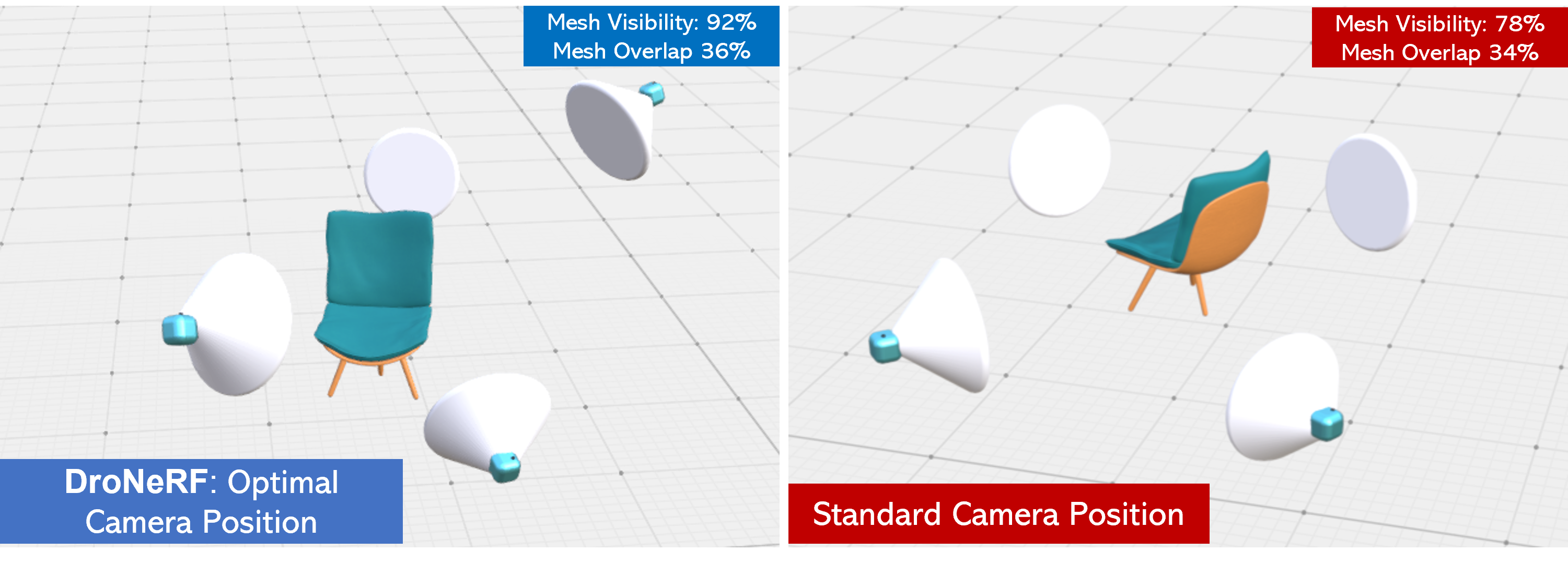}}%
  \caption{\textit{Simulated positioning of cameras for calculating mesh visibility and mesh overlap. Compared to standard circular pattern poses DroNeRF optimization has almost 15\% higher mesh visibility as it is optimized based on the object geometry.}}
  \label{fig:framework}
\end{figure}

\textbf{The key contributions of our work can be summarized as follows:}
\begin{itemize}
    \item We design an end-to-end pipeline, \textbf{\textit{DroNeRF}}, for optimized monocular camera positioning using multi-agent drones for fast 3D scene reconstruction.
    \item Our work builds reliable radiance fields from images streamed by drones around the object without requiring external localization.
    \item \textbf{\textit{DroNeRF}} outperforms the configuration with non-optimized drone positions. We compare the 3D reconstruction for different objects in both configurations.
\end{itemize}

\noindent For the rest of the paper, we summarize some related work in Section 2. Section 3 describes the overview and methodology of the system in detail. In Section 4, we discuss evaluation metrics, results, and analysis. Finally, Section 5 discusses future directions and limitations and concludes our approach.

\section{RELATED WORK}

We review the literature on two different lines of work, optimal camera placement for a scene and neural radiance fields from monocular images, to build up the algorithm at the intersection of both.

\subsection{Camera Pose Optimization}

Positioning multiple cameras around a scene to capture a large view easily outperforms the single camera system. Tarabanis et al. \cite{Tarabanis1995}, Mavrinac et al. \cite{Mavrinac2012ModelingCI}, Zhao et al. \cite{Zhao2011}, and Fu et al. \cite{Fu2014} have explained the problem of placing different cameras in 3D space from the point of view of video surveillance for a 2D image analysis. Fu et al. \cite{Fu2014} also investigated the challenge of optimal camera placement under different restrictions. As Watras et al. \cite{Watras2018} and Zhao et al. \cite{Zhao2011} describe, the cameras can be situated either to use the least number of cameras for the scene or maximize the extent of the scene captured using a definite number of cameras. In this paper, we attempt to integrate both formulations into one.

Fleishman et al. \cite{Fleishman1999} have contributed significantly to the problem of camera placement and coverage. For any vision system to have state-of-the-art performance, their work emphasizes the extent of the camera field of view, resolution, and volume of the target scene. Assuming the scene's geometry is known prior, they proposed a camera positioning architecture to generate 3D models from 2D images. They also incorporated all the prospective camera poses to include every scene feature.

As described by State et al. \cite{Andrei2006}, an innovative approach is to use a rendered camera in simulation to understand the robustness of every camera. They also described an interactive simulator to design, manipulate and test different multi-camera setup configurations within a 3D environment. It implements a unique form of texture mapping on 3D surfaces to showcase each camera's coverage \cite{Chabra2017, bera2016online}. Additionally, work by \cite{williams2021redirected} optimizes scene understanding by visibility polygons.

\subsection{Neural Radiance Fields}

In recent years, a variety of scene representations have been extensively researched, including meshes \cite{AtlasNet, Pixel2mesh}, point clouds \cite{PointNet}, and implicit functions \cite{OccNets, ConvOccNets}. There have also been many proposals for neural representations, which aim to achieve high-quality rendering or natural signal representation. One notable model is NeRF \cite{NeRF}, which uses radiance fields to address novel view synthesis and has achieved photorealistic quality. Although NeRF has impressive rendering quality, it suffers from slow rendering speeds as it requires many expensive multilayer perceptron (MLP) evaluations to render each pixel. Recent papers propose hybrid representations that combine a fast explicit scene representation with learnable neural network components, resulting in significant speedups over purely implicit methods. Various explicit representations have been explored, including sparse voxels \cite{Plenoxels, DVGO}, triplanes \cite{eg3d, hexplane, kplanes}, and tensor factorizations \cite{tensorf}. 

One downside of the above-mentioned NeRF-based methods is that they rely on the camera poses, which are often unavailable, to train their models. A growing area of research in neural radiance fields aims to eliminate this dependence on imperfectly known camera poses. This independence is especially appealing for creating NeRFs since it eliminates the need for labor-intensive, time-consuming preprocessing steps to obtain the camera poses of the images \cite{Schonberger}.

In terms of multi-agent planning (for swarm navigation), there is a wide body of work ~\cite{wang2023aztr, xia2023intelligent, cheung2018efficient, bera2017sociosense, cheung2016lcrowdv, bera2015efficient}, but for computational overheads, we implement baseline version of 3D-RVO~\cite{snape2010navigating}. Our approach does not assume a specific prior and can be optimized from randomly initialized camera poses. In the traditional SfM or SLAM, the structure and appearance of the scenes are simultaneously reconstructed by optimizing an objective function. Inspired by the joint optimization approach from bundle adjustment in classical SfMs, we propose a structure-aware NeRF without posed camera approach \cite{SaNERF}.

\section{OVERVIEW AND METHODOLOGY}

Finding the optimal camera positions for efficient 3D reconstruction is important in computer vision and computer graphics. The goal is to capture images from different viewpoints to reconstruct a 3D model of the scene.  Our approach involves formulating an optimization problem that finds the camera positions that minimize a specific cost function. The cost function can be based on various factors, such as the quality of the resulting 3D model, the number of cameras required, and the distance between cameras.  The basic idea is to formulate an optimization problem that finds the camera positions that minimize a specific cost function.

As Holt et al. \cite{Holt2007} mentioned, a simple technique implemented here is to position the drones on a circle at a fixed arbitrary radius from the object as a starting point. The drones have been evenly placed since the circle can be divided into equal sections depending on the number of drones used. However, an important consideration is to use a minimum number of drones. Since the drones are placed adjacent to each other, their respective field of view overlaps, which is intended to keep at a minimum. Thus, this system of cameras will cover the object features from all directions with maximum coverage.

\begin{figure}[h]
  \centering
  {\includegraphics[width=1\linewidth]{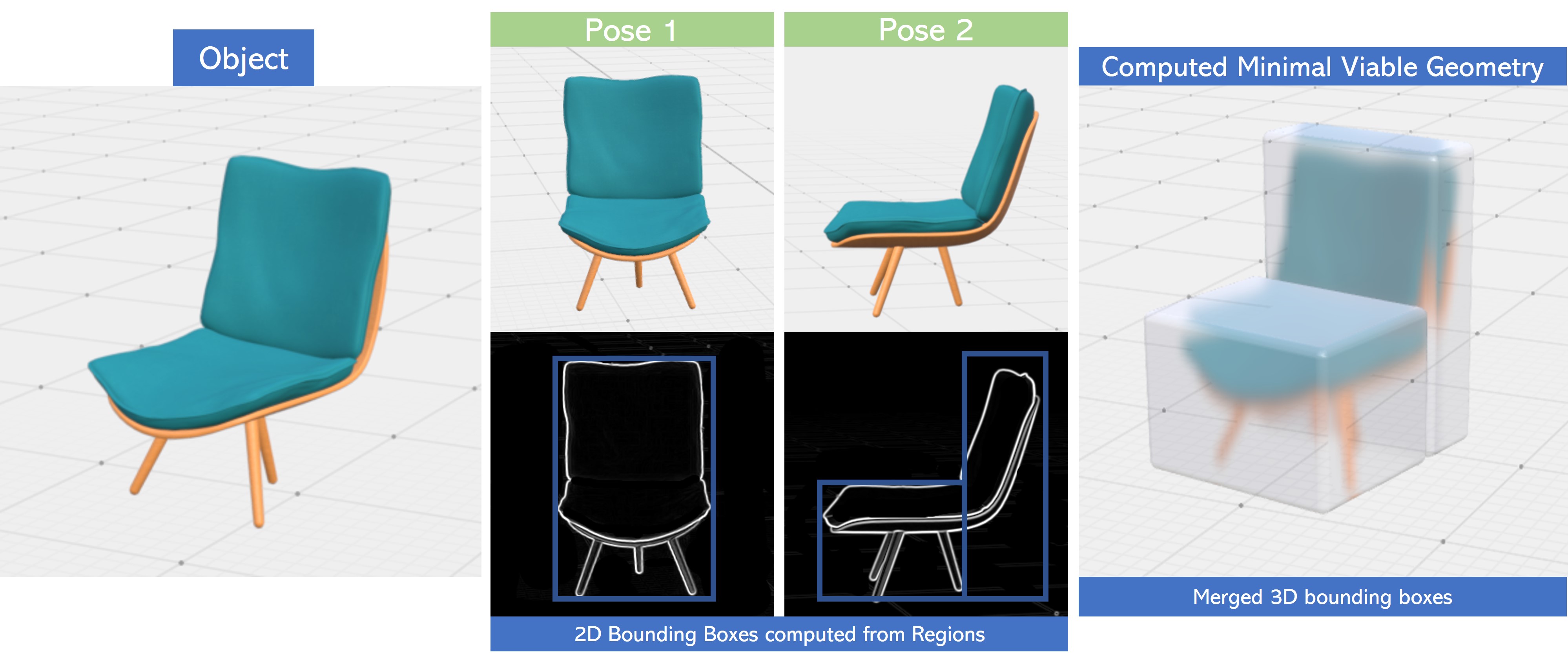}}%
  \caption{\textit{Each camera individually captures the object (as shown on the left) and computes different regions on the entire geometry. Bounding boxes are fitted to the regions respectively (as shown in the center). Finally, all the bounding boxes are merged from each drone view to form a 3D box (as shown on the right).}}
  \label{fig:boundingbox}
\end{figure}

\subsection{Overview}

In order to compute the optimal position of cameras surrounding the object, the system has to understand the scene instantaneously. We intend to construct a 3D model of the object in real-time using a minimal number of images. This reconstruction would be feasible only if the drone prioritizes the 3D geometry of the object to capture principal features instead of the noisy data in the frame. This procedure facilitates efficient and reliable reconstruction in the shortest time frame as the training data gathered would have the most features about the object. Figure \ref{fig:framework} clearly defines our proposed configuration and the conventional uniform arrangement.

\subsection{Drone Configuration Computation}

Inspired by the work of Chabra et al. \cite{Chabra2017}, we use a discrete optimization technique to find the optimal drone configuration for a given scene. We divide the space of possible drone configurations and pose into a set $P$ consisting of $D$ poses, where $D$ is the total number of potential drone poses. We aim to identify a set of optimized drone configurations $O$, where $O$ is a subset of $P$, and $K$ denotes the number of drones used. We initialize $O$ using a greedy algorithm and then refine it using Simulated Annealing (SA). Starting with the initial configuration set produced by the greedy algorithm, the simulated annealing algorithm modifies the camera configurations by either adjusting the poses of the cameras in $O$ or selecting new camera poses from set $P$ in each iteration. If the new configuration's fitness value $F_{fit}$ is better than the best configuration discovered thus far, the algorithm accepts it. We calculate the fitness value based on \cite{Chabra2017}. If the value worsens, the simulated annealing algorithm accepts the new configuration with a probability based on the current temperature and Boltzmann selection \cite{Gonzalez2009}. To avoid local maxima, SA accepts a configuration with a quality metric less than the best-known configuration with a higher probability at the beginning of the run. As the algorithm progresses, the probability of accepting a configuration with a fitness value lower than that of the best-known configuration is decreased until the algorithm only accepts solutions that are better than the current best configuration. Algorithm \ref{alg:cap}, inspired from \cite{Chabra2017}, shows a basic pseudocode for a fitness function to select the optimum pose for a set of drones. It requires camera $C$, Surface $S$, Frames $F$, and $D$ as the input to compute the Fitness function $F_{fit}$.


\begin{algorithm}
\caption{Fitness Function for Multi-camera Pose Optimization}
\label{alg:cap}
\begin{algorithmic}
\Require $C, S, D, F$
\Ensure $F_{fit}$
\State $F_{fit} = 0$
\For {each camera pose $c$ in set $C$}
    \For {each frame $f$ in set $F$}
        \State $s_{near}$ = NULL
        \State $s_{nearDist}$ = $\infty$
        \For {each surface $s$ in set $S$}
            \State $d$ = surfaceDist($c, s, D, f$)
            \If {$d < s_{nearDist}$}
                \State $s_{nearDist} = d$
                \State $s_{near} = s$
            \EndIf
        \EndFor
        \If {$s_{near} \neq$ NULL}
            \State $(x,y)_{expect}$ = expectImgLocation($c, s_{near}$)
            \State $(x,y)_{real}$ = realImgLocation($c, s_{near}, D, f$)
            \State $F'$ = diff($(x,y)_{expect}, (x,y)_{real}$)
            \State $F_{fit} += F'$
        \EndIf
    \EndFor
\EndFor
\State return $F_{fit}$
\end{algorithmic}
\end{algorithm}

The problem here is that we need access to the geometry of the object in question to compute the drone's optimal position. For that, we compute the object's crude, minimal viable geometry as a collection of 3D bounding boxes. In the following subsections, we describe how we compute that efficiently.

\subsection{Seed Geometry Computation}

In order to compute the minimal viable geometry of the scene, we take multiple snapshots from a set of 4 drones surrounding the object. These drones are positioned at 90 degrees orientation to each other. This computes a `region' / edge enclosing the object at the two orthogonal axes. Once we have these collections of these edges, we create multiple 3D bounding boxes to enclose these edges from the two axes.

We define a `region' as an area that can be generalized as a collection of neighboring pixels with common properties like color or intensity to form a colony easily distinguishable from its environment. As compared to edges or corners, regions provide complementary information, which becomes their key identity. 

The Laplacian to the Gaussian kernel can be given as
$$
\nabla^2LoG = \frac{1}{\pi \sigma^4} \left( 1 - \frac{x^2 + y^2}{2\sigma^2} \right)e^{-\frac{x^2 + y^2}{2\sigma^2}} \eqno{(1)}
$$
where $\sigma$ is the standard deviation, and ($x$, $y$) are pixel co-ordinates of Image ${I(x,y)}$.

From (1), it is evident that $LoG$ is dependent on the size of the region structures in the Image $(x, y)$ and the size of the Gaussian kernel $\sigma$ used for smoothing. As the kernel size keeps increasing, it makes the surface smoother, which means the amplitude of the image derivative becomes smaller.

As per Lindeberg \cite{Lindeberg1998}, the derivative decreases exponentially as a function of $\sigma$. To compensate for that, $LoG$ needs to be normalized with $\sigma^2$. Apart from that, to capture regions of all different sizes in the image domain, a multi-scale approach is necessary that automatically scales the selection with different values of $\sigma$. Hence, the scale-normalized Laplacian operator is given as
$$
\nabla^{2}_{norm}LoG = \frac{1}{\pi \sigma^2} \left( 1 - \frac{x^2 + y^2}{2\sigma^2} \right)e^{-\frac{x^2 + y^2}{2\sigma^2}} \eqno{(2)}
$$

Finally, using (2), the normalized Laplacian of Gaussian convolved with the Image $I(x,y)$ is given as
$$
(\hat {x}, \hat {y}; \hat {\sigma}) = argmax_{(x, y; \sigma)}(\nabla^{2}_{norm}LoG * I(x, y)) \eqno{(3)}
$$
where ($\hat {x}, \hat {y}$) gives the position of the region and $\hat {\sigma}$ gives the size of the region.

As per Lindeberg \cite{Lindeberg1998}, the region is found at the maxima, which are the points that are simultaneously local maxima of ${\nabla^{2}_{norm}LoG}$ with respect to both space and scale.

\subsection{Bounding Box Estimation}

The $LoG$, as described, identifies regions of different sizes. However, only the largest region is taken into consideration since it is assumed that the target object is of a relatively larger size compared to the background in terms of pixel coordinates. The simplest way to have a proper estimate is to compute a bounding box around the largest region using OpenCV's boundingRect function. This estimates the straight bounding box from the region instead of the rotated one as the drone won't be able to hold that position, and it will introduce roll \& yaw error in the dataset. Thus, the resultant bounding box will occupy one of the largest areas of the Image. From (3), finding the region with the maximum region size $\hat {\sigma}$ from all $(\hat {x}, \hat {y})$ positions
$$
B_{box} = boundingRect(max(\hat {\sigma}_{(\hat {x}, \hat {y})}))
$$

where $B_{box}$ equates to $(x, y, w, h)$ of the box, which provides $C_{box}$, centroid of the bounding box. Figure \ref{fig:boundingbox} gives the overview of deriving the bounding box from regions.

The centroid $C_{box}$ is utilized to control the position of the drone using the \textit{PID} controller. Since the drone does not have any extra sensors or any external positioning system for feedback, this control strategy would be an open-loop system. However, the error estimation from the geometry of the object makes it function as a closed-loop system.

The \textit{PID} with respect to $C_{box}$ aligns the drone by translating it to the desired location. However, the drone is still out of orientation. Yaw control of the drone is of utmost importance; otherwise, the data captured would be completely misaligned, resulting in poor model formation. To solve this problem, inspired by the work of Chabra et al. \cite{Chabra2017}, we account for the orientation of the centroid of the bounding box $C_{box}$ with respect to monocular camera drone $C$ as
$$
O(C, C_{box}) = \frac{\overrightarrow{C} - \overrightarrow{C_{box}}}{||\overrightarrow{C} - \overrightarrow{C_{box}}||}.\overrightarrow{C_{box_n}} \eqno{(4)}
$$

where ${\overrightarrow{C_{box_n}}}$ is normal to the centroid of the bounding box, which is used to control the yaw error estimate.

Using (4) and $C_{box}$, the \textit{PID} output adjusts the $R_i$, $P_i$, $T_i$, and $Y_i$ (\textit{Roll, Pitch, Throttle, and Yaw}) of the drone to reduce the error between centroid $C_{box}$ and image frame center $(x_{cent}, y_{cent})$. Here, $R_i$, and $T_i$ reduces the error in \textit{X} and \textit{Z} direction, while $P_i$ reduces the error in \textit{Y} direction to maximize object area. Similarly, $Y_i$ reduces the error in $\theta$ plane. Each drone's position is constantly monitored independently to calculate the error and update the PID output in real time to maintain the desired setpoint accurately.

After completing the initial iteration, the drones undergo a pose change by moving in a predefined circular pattern by a fixed amount. Thereafter, the proposed algorithm only fine-tunes the drone pose as described above. As a result, there is no longer a requirement for an external localization system. This entire process is repeated 5 times to capture a total of only 20 images for each object.



\subsection{Scene Reconstruction}
Our goal is to reconstruct the 3D scene in near real-time. To this end, we adopt the multiresolution hash encoding method from Instant-NGP \cite{instant-ngp}.

Like NeRF, we express the continuous scene as a 5D vector-value function, where the input variables consist of a 3D coordinate $(x, y, z)$ and a 2D viewing direction $(\theta, \phi)$. The output variables comprise of the emitted color $(r, g, b)$ and volume density $\sigma$. In their original work, Mildenhall et al. \cite{NeRF} suggested that using positional encoding to encode the 5D input vectors facilitates the model to capture data that contains high-frequency variation. 

Unlike NeRF, Instant-NGP uses a hash-based encoding scheme that results in much faster convergence. The idea is that given a neural network $m$, it contains not only the trainable weights $\textbf{w}$, but also the trainable encoding parameters $\theta$ that encode the input $\textbf{y} = \textnormal{encode}(\textbf{x}, \theta)$. These parameters are organized into $L$ levels, in which each level is associated with $T$ feature vectors.

The hash encoding process can be described as follows:
\begin{itemize}
    \item Identify the neighboring voxels at each of the $L$ resolution levels for the given input coordinate $\textbf{x}$, and assign indices to their corners using a hash function.
    \item Retrieve the corresponding F-dimensional feature vectors from the hash tables $\theta_l$ for all resulting corner indices.
    \item Perform a linear interpolation of the feature vectors based on the relative position of $\textbf{x}$ within the respective $l^{th}$ voxel.
    \item Concatenate the result of each level, along with any auxiliary inputs $\xi \in R^E$ (view direction, material parameters, etc.), to create the encoded input vector $\textbf{y} \in R^{LF+E}$.
    \item This vector $\textbf{y}$ will then be fed to a multilayer perceptron (MLP) to train the model. The output will be the emitted color $\textbf{c}$ and the volume density $\sigma$ like the original NeRF model.
\end{itemize}

The grid resolution is chosen using geometric progression between the coarsest and finest resolutions. To have a complete overview of the hash encoding technique, the original paper by M\"uller et al. \cite{instant-ngp} can be referred to for more detailed explanations. We adopt this multiresolution hash encoding approach in our pipeline. 

\subsection{Experimentation}

The drone involved with this work is the Ryze Robotics Tello EDU version, a small, lightweight, and inexpensive programmable drone. We utilized four such drones to create a swarm. Tello has a 720p, 5MP camera with 960 x 720 resolution with 30 FPS and an approximate flight time of 13 minutes. The bottom of the drone has a Vision Positioning System (VPS) consisting of a camera and a 3D infrared module. This system helps maintain the drone’s current position for precise hovering.

Tello has its own Software Development Kit (SDK) with a list of commands for access to flying and configuration settings. These drones have a built-in hotspot that the phone or PC can connect to send flight commands through their app or any open-source Python modules over the UDP socket communication protocol.

There are two different modes of Tello EDU:
\begin{itemize}
    \item \textbf{Access Point Mode:} The drone acts as an instant access point for a direct one-to-one connection with any device to receive commands and stream the camera view to the device. In this mode, all drones have the same default IP - 192.168.10.1.
    \item \textbf{Station Mode:} The station mode is a mesh connection where multiple drones connect simultaneously to the same access point. Thus a central PC connected to the same access point can send commands to all drones simultaneously, enabling a swarm.
\end{itemize}

Due to hardware limitations, camera streaming is not supported in station mode, as all drones stream camera frames on the same port address by default. Since they are all connected to the same network, there is interference between different streams. To address this issue, we individually connect all drones in access point mode to different USB WiFi adaptors. Further, all the WiFi adaptors are connected to the same WiFi network as before to form a mesh to which the central PC connects. This new setup facilitates swarm formation as well as individual camera streaming. As mentioned, before takeoff, these drones are placed around the target object in a symmetrical configuration from every direction to cover all faces of the object.

We performed all our experiments on Ubuntu 22.04.2, NVIDIA GeForce RTX 4090 12 GB of GDDR6X Machine as our central PC to receive individual camera streams, run the optimization algorithm, and for the NeRF reconstruction. Our proposed module runs in real-time on four image streams of 960x720 resolution. This efficiency is due to parallelizing the optimizing algorithm for individual drones. Every drone took six images each, resulting in 24 images for the reconstruction from individually optimized drone poses. The same steps are repeated for the standard uniform drone configuration as well.

\section{RESULTS AND ANALYSIS}

\subsection{Evaluation Metrics}

For photometric accuracy, we use the Peak Signal-to-Noise Ratio (PSNR) and Structural Similarity Index Measure (SSIM) to evaluate the quality of the rendered images for the NeRF generated using conventional 360-degree poses and the optimized poses using our method.

PSNR measures the difference between the original and reconstructed images. It computes the ratio between the maximum possible value of a pixel and the mean squared error (MSE) between the original and reconstructed images. Higher PSNR scores indicate lower distortion or noise in the reconstructed Image.

SSIM, on the other hand, evaluates the structural similarity between the original and reconstructed images. It considers the images' luminance, contrast, and structure and computes a score ranging from -1 to 1, with 1 indicating perfect similarity. Unlike PSNR, SSIM is designed to be more perceptually accurate, as it incorporates the idea that the human visual system is more sensitive to changes in structural information than pixel values.

\subsection{Qualitative and Quantitative Results}
We compare our approach with a baseline method of drone placement, in which the drones are uniformly placed around an object of interest. We conduct our experiments for four scenes: \textit{human, guitar, robot dog}, and \textit{recycle bin}. We reconstruct 2 NeRF models for each scene: one from a regular drone setup and one from our optimized setup. Figure~\ref{fig:dronemain} represents the qualitative results of our approach in comparison to a baseline method. The images on the left are rendered from NeRF model constructed using regular drone placement. The images on the rights are rendered from the NeRF model of the same scenes but reconstructed using DroNeRF optimization. The NeRF models constructed by our method have better quality than those constructed by the baseline. The difference is that the uniform drone setup does not consider the advantageous camera positions, resulting in the loss of information. DroNeRF addresses this issue by detecting the largest region of interest, computing the corresponding bounding box, then adjusting the drones to the desired locations accordingly. This allows the drones to capture images containing the central object's most important details, resulting in a better NeRF model. We show more results on our website: {\small \href{https://ideas.cs.purdue.edu/research/dronerf}{ideas.cs.purdue.edu/research/dronerf}}.

Table \ref{tab:res} compares the quantitative results of our model with the baseline. To compute the PSNR and SSIM scores, we split the drone-captured images into two sets: a training set and a test set. The training images are used to optimize the NeRF model. The test set contains unseen images with novel viewing angles (or camera poses). These images are then used together with the rendered images from the optimized NeRF model at the same camera angles to compute the PSNR and SSIM scores. 

\begin{table}[h]
    \centering
    \vspace{0.25cm}
    \begin{tabular}{|c|c|c|c|c|}
        \toprule
        \hline
          & \multicolumn{2}{c|}{\textbf{PSNR}}           & \multicolumn{2}{c|}{\textbf{SSIM}} \\
          \hline
        Scenes      & Standard & Optimized & Standard & Optimized \\
        \hline
        Human       & 11.30    & 18.93     & 0.43     & 0.66     \\
        \hline
        Guitar      & 8.89     & 17.67     & 0.29     & 0.56     \\
        \hline
        Robot dog   & 10.18    & 22.72     & 0.40     & 0.70     \\
        \hline
        Recycle bin & 9.26     & 20.13     & 0.34     & 0.67     \\
        \hline
    \end{tabular}
    \caption{\textit{Photometric results for the standard (uniform placement) and optimized (DroNeRF) drone positions for different scenes.}}
    \label{tab:res}
\end{table}

The results show that for all four scenes, the optimized drone positions using DroNeRF achieve higher PSNR and SSIM scores than the standard uniform placement. Specifically, for \textit{Robot dog} scene, the baseline approach achieves a PSNR of 10.18 and an SSIM of 0.40, while DroNeRF achieves a PSNR of 22.72 (123\% increase) and an SSIM of 0.70 (75\% increase). For other scenes, DroNeRF gains at least a 68\% increase in PSNR and a 53\% increase in SSIM (for \textit{human scene}) in comparison to the baseline approach. Overall, the results suggest that the optimized drone positions using DroNeRF can lead to higher image quality compared to the standard uniform placement for photometric reconstruction tasks. However, it is important to note that the evaluation is done on a limited number of scenes, and further evaluations are needed to generalize the findings.

\section{CONCLUSION AND LIMITATIONS}

We propose a new method called DroNeRF for optimizing the placement of drones in photometric 3D reconstruction tasks. The proposed method takes into account the occlusion and visibility of scene surfaces from different drone viewpoints to improve the accuracy of the reconstructed scene. The unique camera positioning significantly impacts the quality of the 3D model, especially for asymmetric objects. By prioritizing the object's geometry, we demonstrate that DroNeRF outperforms the uniform placement of drones in terms of photometric accuracy, as measured by PSNR and SSIM. Moreover, the results show that the proposed method is effective across different scenes and can achieve significant improvements in photometric quality. Overall, we present a promising approach to improving the quality of photometric 3D reconstruction using drone imaging, which could have important applications in fields such as archaeology, architecture, and environmental monitoring.

\textbf{Limitations:} The algorithm relies solely on object geometry to position the drones. This may result in inaccuracies in the pose estimation, especially in dynamic or changing environments. Additionally, the algorithm assumes that the object of interest is stationary and does not consider moving objects. This could result in a noisy 3D model when the scene contains the motions of some objects. This opens opportunities for future improvements.









{\small
\bibliographystyle{IEEEtran}
\bibliography{refs}
}

\end{document}